\documentclass{llncs}
 \pdfoutput=1
\usepackage{graphicx}
\usepackage{amsmath,amssymb} 
\usepackage{color}
\usepackage[width=122mm,left=12mm,paperwidth=146mm,height=193mm,top=12mm,paperheight=217mm]{geometry}
\usepackage{url}
\begin{document}

\newcommand{\pic}[2]{
 \includegraphics[width=#2\textwidth]{#1}
}

\newcommand{\pf}{\noindent \textbf{Proof:} }

\newcommand{\etal}{\emph{et al.}~}

\newcommand{\labeq}[1]{\label{eq:#1}}

\newcommand{\refeq}[1]{(\ref{eq:#1})}

\newcommand{\para}[1]{\noindent \textbf{#1} \,\,}

\newcommand{\paraEmph}[1]{\noindent \emph{#1 --} }

\newcommand{\paraEm}[1]{\noindent \emph{#1} \,\,}

\newcommand{\hs}{\hspace{-0.35cm}}

\newcommand{\vs}{\vspace{-0.18in}}

\newcommand{\picwh}[3]{
 \includegraphics[width=#2\textwidth,height=#3\textheight]{#1}
}

\pagestyle{headings}
\mainmatter
\def\ECCV14SubNumber{269}  

\title{SRA: Fast Removal of General Multipath for ToF Sensors} 



\author{Daniel Freedman*, Eyal Krupka*, Yoni Smolin*, Ido Leichter*, Mirko Schmidt$^\dagger$}
\institute{*Microsoft Research \quad $^\dagger$Microsoft Corporation}

\maketitle

\begin{abstract}
A major issue with Time of Flight sensors is the presence of multipath interference.  We present Sparse Reflections Analysis (SRA), an algorithm for removing this interference which has two main advantages.  First, it allows for very general forms of multipath, including interference with three or more paths, diffuse multipath resulting from Lambertian surfaces, and combinations thereof.  SRA removes this general multipath with robust techniques based on $L_1$ optimization.  Second, due to a novel dimension reduction, we are able to produce a very fast version of SRA, which is able to run at frame rate.  Experimental results on both synthetic data with ground truth, as well as real images of challenging scenes, validate the approach.
\end{abstract}

\section{Introduction}
\label{sec:Introduction}

The field of depth sensing has attracted much attention over the last few years.  By providing direct access to three-dimensional information, depth sensors make many computer vision tasks considerably easier.  Examples include object tracking and recognition, human activity analysis, hand gesture analysis, and indoor 3D mapping; see the comprehensive review in \cite{han2013enhanced}.

Amongst depth sensing technologies, Time of Flight (ToF) imaging has recently shown a lot of promise.  A phase modulated ToF sensor works by computing the time -- measured as a phase-shift -- it takes a ray of light to bounce off a surface and return to the sensor.  ToF sensors are generally able to achieve very high accuracy, and -- since they use light in the infrared spectrum -- to operate in low illumination settings.

The main issue with ToF sensors is that they suffer from \emph{multipath interference} (henceforth simply ``multipath'').  Since rays of light are being sent out for each pixel, and since light can reflect off surfaces in myriad ways, a particular pixel may receive photons originally sent out for other pixels as well.  An illustration is given in Figure \ref{fig:MPDiagrams}.  Significant multipath is observed, for example, in scenes with shiny or specular-like floors.

\begin{figure}[!t]
\centerline{\hbox{
\pic{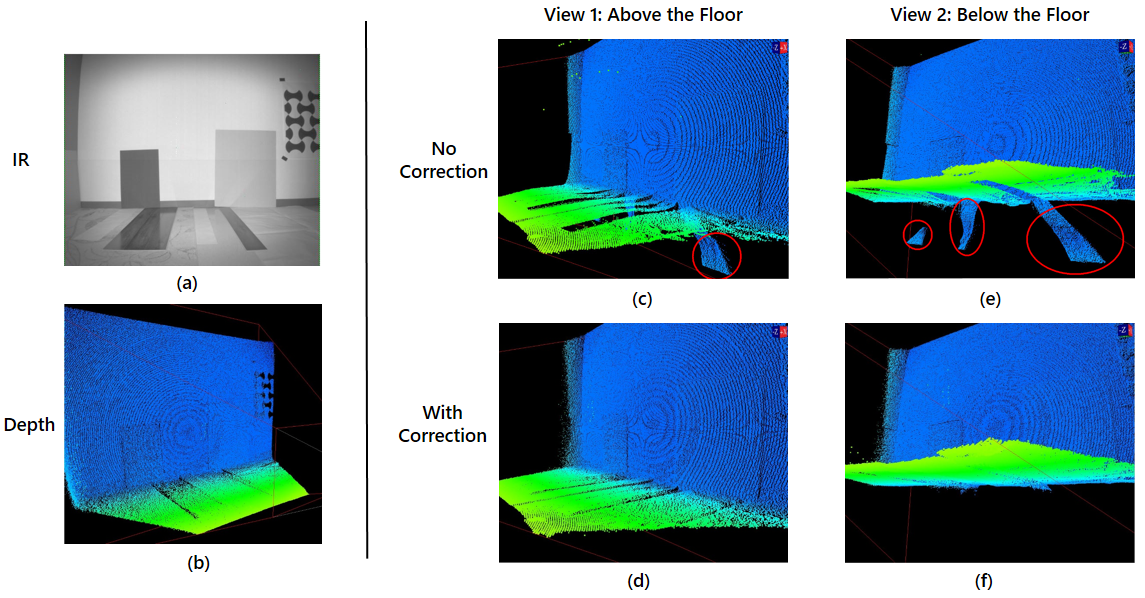}{0.8}  
}}
\caption{Effect of Multipath and its Removal.  Multipath is caused by specular floor materials.  (a) IR image.  (b) Depth reconstruction using our proposed SRA algorithm, rendered from a side-on viewpoint; the floor is shown in green, the wall in blue.  (c, e) Results of not correcting for multipath, shown from two viewpoints -- above and below the floor; gross errors are circled in red.  (d, f) Output of our proposed SRA algorithm.\vs}
\label{fig:Teaser}
\end{figure}

The key problem is that multipath results in corrupted sensor measurements.  These corruptions do not look like ordinary noise, and can be quite large, resulting in highly inaccurate depth estimates; see Figure \ref{fig:Teaser}.  Removing the effect of multipath is therefore a crucial component for ToF systems.

\subsection{Contributions}

Our work addresses two areas in which we improve on the state-of-the-art:

\para{1. More General Multipath}  As we will discuss more explicitly in Section \ref{sec:prior}, prior work falls into two categories.  The first class of algorithms focuses on the case with diffuse multipath, arising from Lambertian surfaces.  The second class of algorithms focuses on the case of ``two-path'' multipath, which arises from specular surfaces.  But multipath can often be more general than this: specular multipath with more than two paths is possible, as are combinations of diffuse and specular multipath.  It is important to note that these are not merely theoretical possibilities; we will show examples of very simple scene geometries which lead to these kinds of combinations.  SRA tackles general multipath, which can include both diffuse and specular multipath, including cases of three or more paths.

\para{2. Speed} Prior work targeting diffuse multipath is very slow, typically requiring a few minutes per frame.  By contrast, SRA is able to run at frame-rate, i.e. 30 fps.  Thus, SRA is the first algorithm, to our knowledge, which allows for diffuse multipath and yet runs in real time.\footnote{Note that prior work targeting ``two-path'' specular multipath does run in real time.}

\subsection{Paper Outline}

\para{1. More General Multipath} In Section \ref{sec:SRA}, we begin by developing a rigorous model which captures a rich class of multipath phenomena.  This model is based on the idea of a sparse, or compressible, backscattering function.  We then show how to formalize the multipath estimation problem as an $L_1$ optimization problem; this is the basis for our Sparse Reflections Analysis (SRA) approach.  SRA is posed in such a way as to admit the computation of a global optimum, which is crucial for the robust cancellation of general multipath even in the presence of considerable measurement noise.

\para{2. Speed} Solving the SRA optimization is relatively fast on a per pixel basis, but would be too slow for real-time performance.  Hence, in Section \ref{sec:dimension} we introduce a novel dimension reduction which allows for a look-up table based approach.  This gives extremely fast performance in practice.

\para{3. Experimental Validation} In Section \ref{sec:experimental} we validate SRA in three ways.  First, we show that in scenes with general multipath -- both specular multipath with more than two paths and combinations of diffuse and specular multipath -- SRA produces accurate depth reconstruction, which competing methods cannot.  Second, we show that in the case of more standard two-path multipath, SRA outperforms the state-of-the-art.  Third, we show the performance of SRA on several challenging images produced by Microsoft's Xbox One sensor \cite{xbox2014xboxone}.

\section{Prior Work}
\label{sec:prior}

Earlier work proposed removing multipath by using additional sensors based on structured light \cite{falie2008distance,falie2008further}, while more recent work has attempted to remove the multipath directly from the sensor measurement itself.  A summary of this more recent work is given in the following table.

\noindent
\begin{tabular}{|p{0.25\linewidth}|p{0.2\linewidth}|p{0.23\linewidth}|p{0.28\linewidth}|}
\hline
\textbf{Paper} & \textbf{Multipath Type} & \textbf{Running Time Per Frame}	& \textbf{Other Constraints} \\
\hline
Fuchs \cite{fuchs2010multipath} & Diffuse only & 10 minutes & \\
\hline
Fuchs \etal \cite{fuchs2013compensation} & Diffuse only & 60-150 seconds & \\
\hline
Jim{\'e}nez \etal \cite{jimenez2012modelling} & Diffuse only & ``Several minutes'' & \\
\hline
Dorrington \etal \cite{dorrington2011separating} & Two-path only & No information & \\
\hline
Godbaz \etal \cite{godbaz2012closed} & Two-path only* (see text)
& No information & Requires 3 or 4 modulation frequencies \\
\hline
Kirmani \etal \cite{kirmani2013spumic} & Two-path only & ``Implementable in real-time'' & Requires 5 modulation frequencies \\
\hline
\end{tabular}

The works of Fuchs \cite{fuchs2010multipath}, Fuchs \etal \cite{fuchs2013compensation}, and Jim{\'e}nez \etal \cite{jimenez2012modelling} all model only diffuse multipath (arising from Lambertian surfaces).  \cite{fuchs2010multipath} estimates the scene by a point cloud and updates the multipath from all pixels to all pixels by ray tracing.   A single pass approximation is performed, whose complexity is quadratic in the number of pixels.  \cite{fuchs2013compensation} is a generalization of \cite{fuchs2010multipath} to a spatially varying, unknown reflection coefficient. It requires an iterative solution consisting of multiple passes.  \cite{jimenez2012modelling} performs a somewhat different iterative optimization of a global function involving scene reconstruction and ray tracing.  As is noted in the table, none of these methods are close to real-time, requiring anywhere between 1-10 minutes of processing per frame.

The works of Dorrington \etal \cite{dorrington2011separating}, Godbaz \etal \cite{godbaz2012closed}, and Kirmani \etal \cite{kirmani2013spumic} all model two-path multipath, arising from specular surfaces.  All of these methods work on a per pixel basis, using either closed form solutions \cite{godbaz2012closed,kirmani2013spumic} or optimizations \cite{dorrington2011separating}.  Thus, while they do not report on their running times explicitly, it is reasonable to expect that they may be close to real-time.  \cite{kirmani2013spumic} requires 5 modulation frequencies; one of the two methods presented in \cite{godbaz2012closed}, which is based on a Cauchy distribution approximation to the backscattering of a single return, requires 4.  Several commercial ToF sensors, such as the Xbox One, use only 3 modulation frequencies; thus, these methods are rendered impracticable for such sensors.  By contrast, the second method presented in \cite{godbaz2012closed}, which uses a more standard delta-function approximation to a single return, only requires 3 modulation frequencies.  Given the additional fact that this method runs in or near real-time, it is therefore our nearest competitor.

\section{Sparse Reflections Analysis}
\label{sec:SRA}

\subsection{The Multipath Representation}
\label{sec:representation}

\para{The ToF Measurement} We begin by describing the vector which is measured by a ToF sensor.  For a given pixel, the sensor emits infra-red (IR) light modulated by several frequencies.  The light bounces off a surface in the scene, and some of the light (depending on the reflectivity and orientation of the surface) is returned to the detector.  For each of $m$ modulation frequencies, this light is then integrated against sinusoids with the same frequency, such that the phase of the measurement $v$ is based on the distance to the surface:
\begin{equation}
v \in \mathbb{C}^m, \text{ with } v_k = x e^{2 \pi i d / \lambda_k}, \quad k = 1, \dots, m
\labeq{noMultipath}
\end{equation}
where $d$ is the distance to the surface, $\lambda_k = c/2f_k$ is half of the wavelength corresponding to the $k^{th}$ modulation frequency $f_k$, and $x$ is a real scalar corresponding to the strength of the signal received.  A typical choice for the number of frequencies is $m = 3$; this is generally sufficient to prevent aliasing effects.

\begin{figure}[!t]
\centerline{\hbox{
\begin{tabular}{cc}
\pic{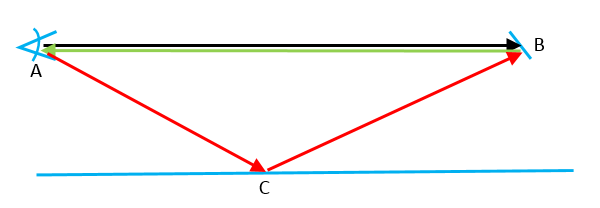}{0.23} & \pic{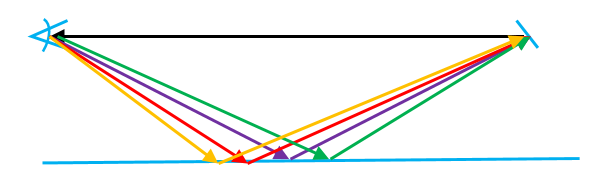}{0.23} \\
(a) & (b)
\end{tabular} }}
\caption{Illustration of Multipath.  (a) Two-Path Multipath. The camera (A) and surface (B) are shown in blue, and light rays in other colours.  The correct path is A-B-A; the second, incorrect path is A-C-B-A.  (b) Diffuse Multipath.  This results from small reflections from many nearby points, shown here as four colored paths.  In fact, there are an infinite number of such points.\vs}
\label{fig:MPDiagrams}
\end{figure}

\para{Two Paths} Equation \refeq{noMultipath} assumes that there is no multipath in the scene.  If there is single extra path (the ``two-path'' scenario), as shown in Figure \ref{fig:MPDiagrams}(a), then the above equation is modified to
\begin{equation}
v_k = x_1 e^{2 \pi i d_1 / \lambda_k} + x_2 e^{2 \pi i d_2 / \lambda_k}
\labeq{twoPath}
\end{equation}
where $d_1$ and $d_2$ are the distances of the two paths, and $x_1$ and $x_2$ give the strengths of the two paths.  If $d_1 < d_2$, then $d_1$ is the true distance and $d_2$ is the multipath component; and the ratio $x_2 / x_1$ gives the strength of the multipath.

\para{Diffuse Multipath} An ideal Lambertian surface receives (a finite amount of) light from a given direction, and reflects infinitesimal amounts in all directions.  So a given point on the Lambertian surface can be responsible for at most an infinitesimal amount of multipath.  However, if an infinite number of nearby points all reflect infinitesimal amounts, then the result can be finite.  This is shown in Figure \ref{fig:MPDiagrams}(b).

In fact, the measurements under diffuse multipath will appear as follows:
\begin{equation}
v_k = x_1 e^{2 \pi i d_1 / \lambda_k} + \int_{d_1+\Delta}^\infty x_L(c) e^{2 \pi i c / \lambda_k} dc
\labeq{lambertian}
\end{equation}
for some $\Delta \ge 0$. The shape of $x_L(c)$ can be determined by looking at simulations of diffuse multipath, and turns out to be well approximated by $x_L(c) \approx A c^\alpha e^{-\beta c}$, where $\alpha$ and $\beta$ depend on the geometry of the underlying scene.

\para{General Multipath} Of course, one can have more general multipath, including: three or more paths; diffuse plus two-path; and so on.  Examining Equations \refeq{twoPath} and \refeq{lambertian}, one can generalize as $v_k = \int_0^\infty x(c) e^{2 \pi i c / \lambda_k} dc$. In what follows, we will use a discrete version of the foregoing, namely
\begin{equation}
v_k = \sum_{j=1}^n x_j e^{2 \pi i d_j / \lambda_k}
\labeq{mpGeneral}
\end{equation}
where $d_j$ is over the relevant interval.  In typical examples, we take the range of object distances to be 20 cm to 450 cm, with increments of 1 cm.  Thus, in this case $n = 431$.

It will be useful to rewrite Equation \refeq{mpGeneral} in vector-matrix form as $v = \Phi x$, where $\Phi \in \mathbb{C}^{m \times n}$ and $x \in \mathbb{R}^n$.  The vector $x$ is referred to as the \textbf{backscattering}.  We then make a further modification to turn the complex measurements into real ones, by stacking.  In particular, we stack the real part of $v$ on top of the complex part, and abuse notation by denoting the $2m$-dimensional result also as $v$.  We do the same with $\Phi$, leading to a $2m \times n$ real matrix given by
\[
\Phi_{kj} = 
\begin{cases}
\cos(2 \pi d_j / \lambda_k) & \text{if } k = 1, \dots, m \\
\sin(2 \pi d_j / \lambda_{k-m}) & \text{if } k = m+1, \dots, 2m
\end{cases}
\]
Then we may still write
\begin{equation}
v = \Phi x
\labeq{mpGeneralMatrix}
\end{equation}
but now all quantities are real.

\para{A Characterization of the Backscattering}
Let us now turn to characterizing the types of backscatterings $x$ which are relevant to capturing the multipath phenomenon.

\paraEm{Property 1: Non-Negativity} The first property, which may seem obvious, is that $x$ is non-negative: $x \ge 0$.  That is, there can only be positive or zero returns for any given distance.

\paraEm{Property 2: Compressibility} The second property is more interesting.  We saw that the two-path scenario involved an $x$ which was zero at all indices except for two (corresponding to $d_1$ and $d_2$ in Equation \refeq{twoPath}).  Such an $x$ is sparse.

On the other hand, the diffuse multipath of Equation \refeq{lambertian}, which has the form $x_L(c) \approx A c^\alpha e^{-\beta c}$, is not sparse.  Rather, its discretized version has the following property: when the $x$ coefficients are sorted from greatest to smallest, the resulting vector falls off quickly to $0$.  This property is referred to as compressibility.

Formally, given a vector $x = (x_1, \dots, x_n)$, let $(x_{I(1)}, \dots, x_{I(n)})$ denote the vector sorted in descending order.  Then $x$ is compressible if $x_{I(i)} \le Ri^{-1/r} $ with $r \le 1$.  That is, the sorted entries of $x$ fall off as a power law.

\subsection{The SRA Algorithm}
\label{sec:SRAAlg}

We have represented multipath via the backscattering $x$, and have further characterized the important properties of the backscattering -- non-negativity and compressibility.  We now go on to show how to use this information to cancel the effects of multipath, and hence find a robust and accurate depth estimate from the raw ToF measured vector.

\para{Multiplicity of Solutions} We are given $v$, and we know that a backscattering $x$ has generated $v$; i.e. following Equation \refeq{mpGeneralMatrix}, we have $v = \Phi x$.  Given that $x \in \mathbb{R}^n$ and $v \in \mathbb{R}^{2m}$ where $n \gg 2m$, there are many possible $x$'s which can generate $v$.  But our characterization of the backscattering says that $x$ is non-negative and compressible, which leads to a much more restrictive set of possible backscatterings.

In fact, due to sensor noise we will not have $v = \Phi x$ exactly.  Rather, we may expect that $v = \Phi x + \eta$, where $\eta$ is generally taken to be Gaussian noise, with zero mean and known covariance matrix $C$.\footnote{In fact, due to the physics of the sensor, there is often a shot noise component involved.  We will ignore this consideration, though our method can be adapted to handle it.}

\para{$L_0$ Minimization} Let us suppose, for the moment, that $x$ is \emph{sparse} rather than compressible -- that is, $x$ has a small number of non-zero entries.  The number of non-zero entries of $x$ is often denoted as it's 0-norm, i.e. $\| x \|_0$.  In this case, one would like to solve the following problem:
\begin{equation}
\min_{x \ge 0}\, \| x \|_0 \quad \text{subject to} \quad (\Phi x-v)^T C^{-1} (\Phi x - v) \le \epsilon^2 \| v \|^2
\labeq{L0}
\end{equation}
for some parameter $\epsilon$, which we fix to be $0.05$ in our experiments.  (Note that we have indicated the non-negativity of $x$ under the $\min$ itself.)  That is, we want the sparsest backscattering $x$ which yields the measurement vector $v$, up to some noise tolerance.  Unfortunately, the above problem, which is combinatorial in nature, is NP-hard to solve.

\para{$L_1$ Minimization} However, it turns out that subject to certain conditions on the matrix $\Phi$, solving the problem
\begin{equation}
\min_{x \ge 0}\, \| x \|_1 \quad \text{subject to} \quad (\Phi x-v)^T C^{-1} (\Phi x - v) \le \epsilon^2 \| v \|^2
\labeq{L1}
\end{equation}
will yield a similar solution \cite{candes2005decoding,candes2006stable,donoho2006compressed} to the optimization in \refeq{L0}.  Note that the only difference between the two optimizations is that we have replaced the 0-norm with the 1-norm.  The key implication is that the optimization in \refeq{L1} is \emph{convex}, and hence may be solved in polynomial time.

In fact, the conditions mentioned in \cite{candes2005decoding,candes2006stable,donoho2006compressed}, such as the Restricted Isometry Property, are generally not satisfied by our matrix $\Phi$.  This is due to the redundancy present in nearby columns of $\Phi$, i.e. nearby columns tend to have high inner products with each other.  Nevertheless, we can use the optimization in \refeq{L1}, with the understanding that certain theoretical guarantees given in \cite{candes2005decoding,candes2006stable,donoho2006compressed} do not hold.  Note that this is in the same vein as other computer vision work, such as the celebrated paper by Wright \etal on robust face recognition \cite{wright2009robust}, which used $L_1$ optimization under conditions which differed from those specified in \cite{candes2005decoding,candes2006stable,donoho2006compressed}.

Until now we have been assuming that $x$ is sparse, rather than compressible.  It turns out, however, that even if $x$ is compressible and not sparse, then solving the $L_1$ optimization in \refeq{L1} still yields the correct solution \cite{candes2005decoding,candes2006stable,donoho2006compressed}.

\para{$L_1$ with $L_1$ Constraints} Although the optimization \refeq{L1} is convex, it is a second-order cone program which can be slow to solve in practice.  We therefore make the following modification.  Note that the $L_2$ constraint above may be written $\|C^{-1/2} (\Phi x - v)  \|_2 \le \epsilon \| v \|_2$.  We may consider approximating these constraints by their equivalent $L_1$ constraints, i.e. $\|C^{-1/2} (\Phi x - v)  \|_1 \le \epsilon \| v \|_1$.  In this case, the resulting optimization becomes
\begin{equation}
\min_{x \ge 0}\, \| x \|_1 \quad \text{subject to} \quad \|C^{-1/2} (\Phi x - v)  \|_1 \le \epsilon \| v \|_1
\labeq{L1L1}
\end{equation}

The advantage of this formulation over \refeq{L1} is that it can be recast as a linear program.  As such, it can be solved considerably faster.  To perform the conversion, first notice that since $x \ge 0$, $\| x \|_1 = \sum_{i=1}^n x_i = \mathbf{1}^T x$.  Second, note that the constraint $\|z\|_1 \le \gamma$ for $z \in \mathbb{R}^\ell$ can be converted into the set of linear constraints $Q_\ell z \le \gamma \mathbf{1}$, where $Q_\ell$ is a $2^\ell \times \ell$ matrix, whose rows consist of all elements of the set $\{-1, +1\}^\ell$.  While this might be prohibitive for $\ell$ large, in our case $\ell = 2m$, and we generally have $m = 3$; this leads to $64$ extra constraints, much fewer than the number of non-negativity constraints.  This yields the linear program
\[
\min_{x \ge 0}\, \mathbf{1}^T x \quad \text{subject to} \quad Ax \le b
\]
where $A = Q_{2m} C^{-1/2} \Phi$ and $b = Q_{2m} C^{-1/2} v  + \epsilon \| v \|_1 \mathbf{1}$.

\para{Computing Depth from Backscattering}
The various optimization problems we have just described yield the backscattering $x$.  Of course, in the end our goal is an estimate of the depth; we now explain how to extract the depth from $x$.

The main path must have the shortest distance; this results from the geometry of the imaging process.  Thus, we have simply that the depth corresponds to the first non-zero index of $x$, i.e. the index $i_1(x) \equiv \arg\min_i \{i: x_i > 0 \}$.  Then the depth is just $\delta = d_{i_1(x)}$.  In practice, due to numerical issues there will be many small non-zero elements of $x$.  Thus, we take $i_1(x) \equiv \arg\min_i \{i: x_i > c \max_{i'} x_{i'} \}$ for some small $c$; typically, we use $c = 0.01$.

If we have a reasonably accurate noise model, we can be more sophisticated.  For each peak of the backscattering $x$, we can compute the probability that the peak is generated by noise rather than signal.  Then the probability that the first peak is the true return is just one minus the probability that it is generated by noise; the probability that the second peak is the true return is the probability that the first peak is generated by noise times one minus the probability that the second peak is generated by noise; and so on.  If no return has probability greater than a threshold (e.g. 0.9), we can ``invalidate'' the pixel -- that is, declare that we do not know the true depth.  We will make use of this kind of invalidation in Section \ref{sec:experimental}.

\section{Fast Computation}
\label{sec:dimension}

SRA allows us to compute the backscattering $x$ from a sensor measurement $v$; from the backscattering, one can compute the depth.  The issue that now arises is related to the speed of the computation.  Solving the optimization in \refeq{L1L1} typically requires about 50 milliseconds per instance on a standard CPU.  Given that a ToF image may consist of several hundred thousand pixels, this yields on the order at least an hour per frame.  To achieve a frame rate of 30 Hz, therefore, a radically different approach is needed.  The method we now describe allows for SRA to run at frame rate on ordinary hardware, for images of size $424 \times 512$.

\para{Dimension Reduction: Motivation}
Real-time computation is often aided by performing pre-computation in the form of a look-up table (LUT).  If we construct a LUT directly on the measurement vector $v$, then the table will be $2m$-dimensional, as this is the dimensionality of $v$.  It is easy to see that multiplying $v$ by a scalar does not change the results of any of the SRA optimizations, except to scale $x$ by the same scalar.  Thus, one can easily normalize $v$ so that its $L_2$ norm (or $L_1$ norm) is equal to $1$,  yielding a reduction of a single dimension.  The resulting table will then be $(2m-1)$-dimensional.

Our goal is a further reduction of a single dimension, to $(2m-2)$ dimensions.  Recall that the size of an LUT is \emph{exponential} in its dimension; thus, the reduction of two dimensions reduces the total memory for the LUT by a factor of $L^2$, where each dimension has been discretized into $L$ cells.  This reduction makes the LUT approach feasible in practice.

\para{A Useful Transformation}
Let us return to the complex formulation of the problem; this will make the ensuing discussion easier, though it is not strictly necessary.  Let us define the $m \times m$ complex matrix $F_{s,\Delta}$ by 
\[
F_{s,\Delta} = s \cdot\text{diag}(e^{-2\pi i \Delta / \lambda_1}, \dots, e^{-2\pi i \Delta / \lambda_m})
\]
where $\Delta \in \mathbb{R}$; $s > 0$ is any real positive scalar; and $\text{diag}()$ denotes the diagonal matrix with the specified elements on the diagonal.  Then we have the following theorem, from which we can derive our dimension reduction.
\begin{theorem}
Let $x^*$ be the solution to the optimization \refeq{L1}, and let $x_{s,\Delta}^*$ be the solution to \refeq{L1} with $F_{s,\Delta} v$ replacing $v$ and $F_{s,\Delta} \Phi$ replacing $\Phi$.  Suppose that the covariance $C$ is diagonal, and satisfies $C_{jj} = C_{j+m,j+m}$. Then $x_{s,\Delta}^* = x^*$. 
\label{claim:noisy}
\end{theorem}

The proof is included in the supplementary material.  The implication of Theorem \ref{claim:noisy} is that multiplying both the measurement $v$ and the matrix $\Phi$ by the matrix $F_{s,\Delta}$ does not change the backscattering (assuming the theorem's conditions on the coviarance matrix hold, which is a reasonable model of sensor noise).  Of course, the corresponding range of distances has been shifted by by $-\Delta$, so in extracting the depth from $x_{s,\Delta}^*$, one must add on $\Delta$ afterwards.

Before going on, we note that Theorem \ref{claim:noisy} applies to optimization \refeq{L1} rather than \refeq{L1L1}, which we use in practice.  However, as \refeq{L1L1} is a reasonable approximation to \refeq{L1}, we proceed to use Theorem \ref{claim:noisy} to construct our dimension reduction.

\para{The Canonical Transformation}
The Canonical Transformation is derived from $F_{s,\Delta}$ by a particular choice of $s$ and $\Delta$.  Let $k \in \{1, \dots, m\}$ be a specific frequency index; then Canonical Transformation of $v$, $\rho^{(k)}(v)$, is given by
\[
\rho^{(k)}(v) \equiv F_{s,\Delta} v \quad \text{with} \quad s = \|v\|^{-1}, \,\, \Delta = \lambda_k (\angle v_k / 2\pi)
\]
where $\angle v_k$ denotes the phase of $v_k$, taken to lie in $[0, 2\pi)$.  It is easy to see that $\rho^{(k)}$ has the following property.  The $k^{th}$ element of $\rho^{(k)}(v)$ is real, i.e. has 0 phase.  Furthermore, the $k^{th}$ element of $\rho^{(k)}(v)$ may be found from the other elements of $\rho^{(k)}(v)$ by $\rho_k^{(k)}(v) = \left( 1 - \sum_{k' \neq k} | \rho_{k'}^{(k)}(v) |^2 \right)^{1/2}$.

In other words, in the Canonical Transformation, one of the elements is \emph{redundant}, in that it is completely determined by the other elements.  Hence, this element can be removed without losing information.  Of course, the component is complex, meaning that we have removed two real dimensions, hence enabling the promised dimension reduction from $2m$ to $2m-2$.  A LUT can be built on the remaining $2m-2$ dimensions, simply by discretizing over these dimensions.

Note that having transformed $v$ by $F_{s,\Delta}$ (with $s$ and $\Delta$ given by the Canonical Transformation), we must also apply $F_{s, \Delta}$ to $\Phi$ in order to use Theorem \ref{claim:noisy}.  In fact, this is straightforward: this transformation simply ``shifts'' the columns of $\Phi$.  So if before they represented distances in the range $[D_{min}, D_{max}]$, they now represent distances in the range $[D_{min}-\Delta, D_{max}-\Delta]$.  Rather than actually do the shifting, we simply enlarge this range.  Note that the minimal value that $\Delta$ can take on is $0$, while the maximal value is $\lambda_k$; thus, after the Canonical Transformation the potential distances can now fall between $D_{min} - \lambda_k$ and $D_{max}$.  Thus, the matrix $\Phi$ is now enlarged to have columns corresponding to distances in the range $[D_{min} - \lambda_k, D_{max}]$.  A natural method for choosing $k$ is to keep the above range as small as possible, and hence to choose the $k$ corresponding to the smallest half-wavelength $\lambda_k$.

%

\section{Experiments}
\label{sec:experimental}


\subsection{Running Time}

As one of the main claims of this paper is a fast algorithm, we begin by presenting the speed of the algorithm.  We have benchmarked SRA on images of size $424 \times 512$, which are the standard for the Xbox One sensor.  The code is run on an Intel Core i7 processor, with 4 cores, 8 logical processors, and a clockspeed of 2.4 GHz.  The code runs in 31.2 milliseconds per frame, which is real-time given a frame-rate of 30 fps.  Note that our code is largely unoptimized Matlab (the only optimization we make is to use the CPU's 8 logical processors for parallelization); the speed comes from the  LUT-based approach.  It is to be expected that optimized C code would be even faster.

\subsection{General Multipath: Examples}


\para{Specular Three Path}
We begin by motivating the relevance of general multipath.  Specular multipath with three or more paths results naturally from simple scene geometries.  Suppose that we have the geometry shown in Figure \ref{fig:MPDiagrams}(a), where the object (B) lies on a Lambertian surface and the scene element (C) is taken to be purely specular.  Then it can be shown that by varying the position and normal of the scene element, we can generate any relative amplitudes we wish between the direct (A-B-A) and interfering (A-C-B-A) paths, as well as any pair of path distances; please see the supplementary material  for more details.  Generating three (or more) paths then becomes straightforward, by adding an extra (or multiple extra) specular surfaces to the scene.

%

\begin{figure}[!t]
\centerline{\hbox{
\begin{tabular}{ccccc}
\pic{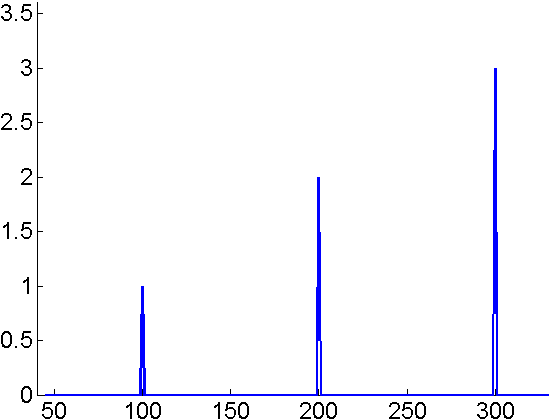}{0.18} & \pic{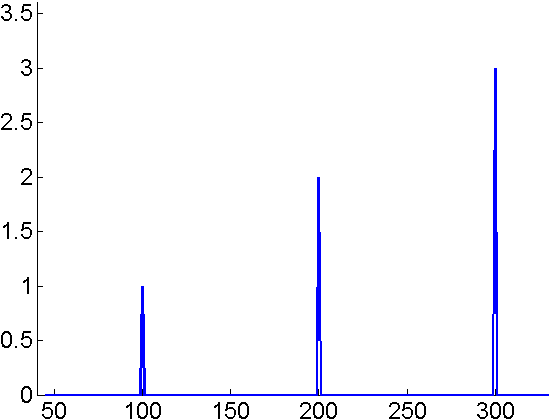}{0.18} & \pic{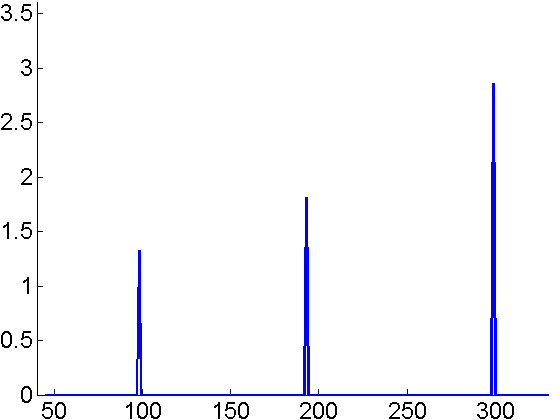}{0.18} & \pic{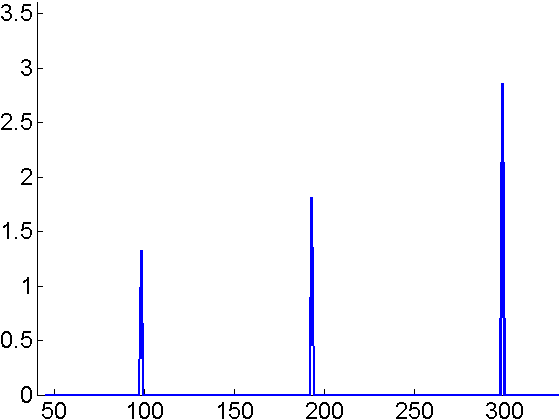}{0.18} & \pic{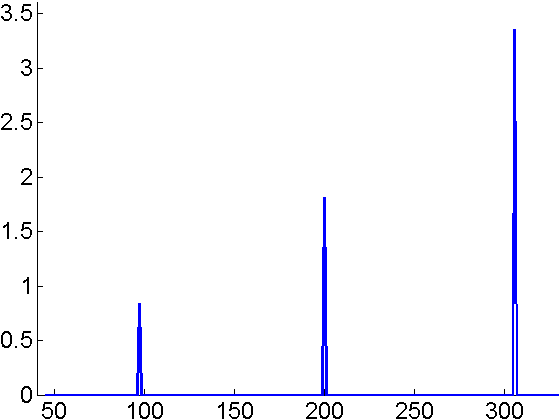}{0.18}
\end{tabular}
}}
\caption{Three-Path Simulation. Left: true backscattering.  Following 4 plots: SRA reconstruction of the backscattering, for SNR$= \infty, 20, 10, 5$.  See discussion in the text.\vs}
\label{fig:ThreePath}
\end{figure}

\begin{figure}[!t]
\centerline{\hbox{
\begin{tabular}{ccc}
\picwh{Ex1True.png}{0.3}{0.1} & \picwh{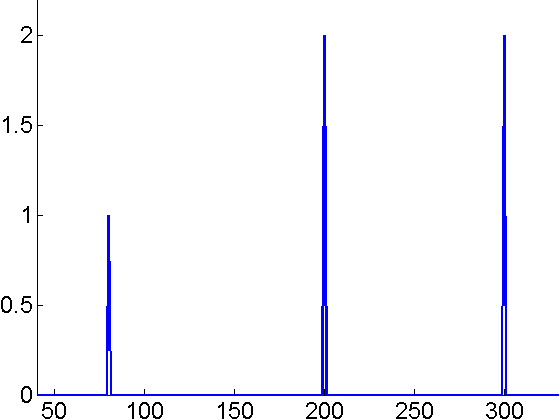}{0.3}{0.1} & \picwh{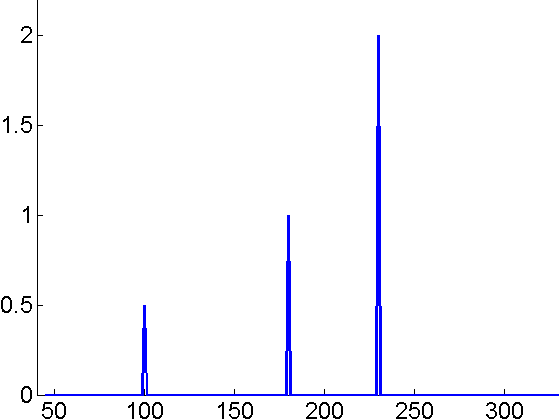}{0.3}{0.1} \\
\begin{tabular}{|c||c|c|c|}
\hline
SNR & SRA & Godbaz & ML \\
\hline
$\infty$ & \textbf{0.0} & 15.0 & 111.0 \\
\hline
20 & \textbf{1.9} & 114.7 & 111.0 \\
\hline
10 & \textbf{3.7} & 119.3 & 111.0 \\
\hline
5 & \textbf{8.1} & 109.2 & 109.0 \\
\hline
\end{tabular}
&
\begin{tabular}{|c||c|c|c|}
\hline
SNR & SRA & Godbaz & ML \\
\hline
$\infty$ & \textbf{0.0} & 8.4 & 29.0 \\
\hline
20 & \textbf{2.1} & 108.4 & 29.0 \\
\hline
10 & \textbf{4.1} & 117.4 & 29.0 \\
\hline
5 & \textbf{8.6} & 111.9 & 30.0 \\
\hline
\end{tabular}
&
\begin{tabular}{|c||c|c|c|}
\hline
SNR & SRA & Godbaz & ML \\
\hline
$\infty$ & \textbf{0.0} & 8.0 & 29.0 \\
\hline
20 & \textbf{8.7} & 91.7 & 29.0 \\
\hline
10 & \textbf{17.4} & 86.2 & 29.0 \\
\hline
5 & 31.7 & 76.4 & \textbf{31.0} \\
\hline
\end{tabular}
\end{tabular}
}}
\caption{Three-Path Simulation.  Each column shows the true backscattering (top), and the median absolute error in cm of three algorithms under various noise levels (bottom).  The best algorithm is indicated in bold. See discussion in the text.\vs}
\label{fig:ThreePathA}
\end{figure}

Figure \ref{fig:ThreePath} shows an example of three path specular interference.  The leftmost plot shows the true backscattering, corresponding to object distances of 100, 200, and 300 cm, with amplitudes in the ratio 1:2:3.  That is, the multipath is $2+3=5$ times stronger than the initial return.  Moving from left to right, the following four plots show the backscattering computed by the SRA algorithm under different levels of noise: SNR$= \infty, 20, 10, 5$.  Note that the backscattering extracted is exactly correct for the case of SNR$=\infty$, and remains fairly close to the true backscattering as the noise level increases.  Indeed, for the highest noise level of SNR$=5$, the peaks have moved to 97, 200, and 306 cm (movements of -3, 0, and 6 cm resp.), which yields an error of 3 cm in depth estimation.  (Note the amplitudes have changed as well, moving less than 20\% in all cases.)

We now move to a more quantitative comparison of SRA with other alternatives in the case of three paths.  As our aim is to show the need to model more general multipath, we run against two strong two-path alternatives: the algorithm of Godbaz \etal \cite{godbaz2012closed} (our most natural competitor, for reasons described in Section \ref{sec:prior}) and the maximum likelihood (ML) two-path solution.  Note that the ML solution is not a practical algorithm, as it requires a slow, exhaustive examination of all pairs of paths; but we include it as it represents the best possible two path solution.  Figure \ref{fig:ThreePathA} shows three separate configurations; the top row shows the true backscatterings, while the tables below show the performance of SRA vs. the two alternatives.  The performance is given by the median absolute error of the depth; as we are adding noise, we average over 1,000 samples.

We note that SRA outperforms the two alternatives, and does so by a wide margin once even a small amount of noise is added.  In the first example, the accuracies of SRA are very high, staying under 4 cm for SNR levels up to 10; the second example is similar.
The third case has been chosen to be more difficult for SRA: the multipath is 6 times stronger than the original path, and the second and third returns are fairly closely spaced.  SRA's errors here are higher, though still considerably lower than the alternatives (except in the case of SNR = 5, where performance is similar to ML).  In all three cases, Godbaz gives reasonable results in the noiseless case, but fails once even a small amount of noise (SNR = 20) is added.  ML is much more resistant to noise, but does not give very high accuracy in any of the examples, regardless of noise level.

\begin{figure}[!t]
\centerline{\hbox{
\begin{tabular}{ccc}
\picwh{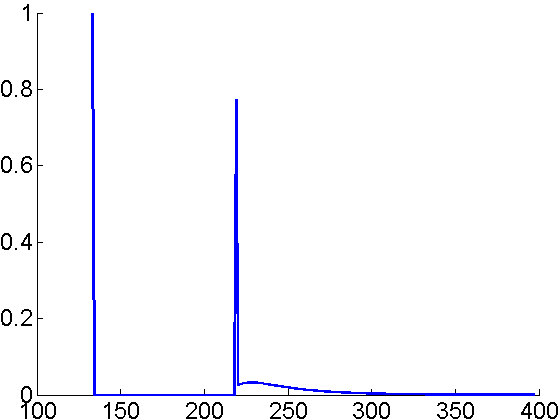}{0.27}{0.08} & \picwh{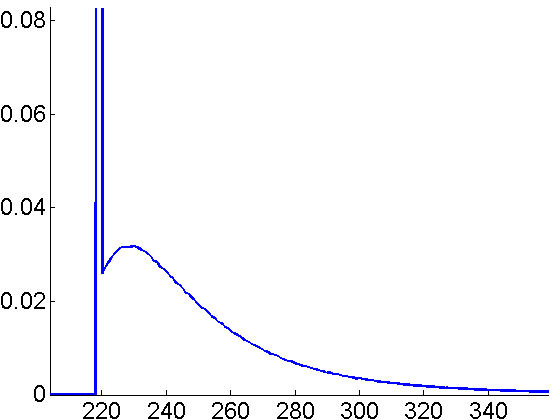}{0.27}{0.08} & \pic{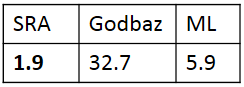}{0.22}
\end{tabular}
}}
\caption{Combined Diffuse and Specular Multipath. Left: true backscattering.  Middle: detail of the backscattering, showing the diffuse part.  Right: Absolute error in cm. \vs}
\label{fig:DiffuseSpecular}
\end{figure}

\para{Combined Diffuse and Specular Multipath}  We now show an example which leads to a combination of diffuse and specular multipath.  The geometry is again simple, and consists of an object and a single plane; both object and plane have both specular and diffuse reflectivity.  The backscattering, which we generate by use of our own light-transport simulator, is shown in Figure \ref{fig:DiffuseSpecular}; note the fact that the backscattering is no longer sparse (but is still compressible).

The depth estimate errors are shown in Figure \ref{fig:DiffuseSpecular}.  The method of Godbaz generates fairly large errors.  ML is considerably better: it turns out that there is a reasonably good two path approximation to the measurement $v$ produced by this backscattering.  SRA produces the lowest error: not surprisingly, allowing for a more complex backscattering -- as SRA does -- leads to the best result.


\subsection{Comprehensive Two-Path Evaluation}

\para{Setup} We have shown the ability of SRA to deal with general multipath.  However, standard two-path interference is a very important case, and we would like to show SRA's capabilities in this regime.  We challenge SRA in two ways: by simulating high multipath, up to a factor of 5 times as high as the direct return; and by simulating high noise regimes.  We again compare against Godbaz \etal \cite{godbaz2012closed}, our most natural competitor (for reasons described in Section \ref{sec:prior}).

In particular, we simulate returns of the form $v_k = x_1 e^{2 \pi i d_1 / \lambda_k} + x_2 e^{2 \pi i d_2 / \lambda_k} + \eta_k$; the noise $\eta_k$ has independent real and imaginary components, and is taken to be Gaussian with variance $\sigma^2$ for each component. There are two critical parameters: (1) Multipath Strength is defined as $x_2 / x_1$, and takes on values in the set $\{$0.6, 1.1,  1.7, 2.2, 2.8, 3.3, 3.9. 4.4, 5.0$\}$.  (2) SNR is defined as $x_1 / \sqrt{6}\sigma$, since there are 6 independent noise components to the measurement.  It takes on values in the set $\{\infty$, 25.5, 12.7, 8.5, 6.4, 5.1, 4.2, 3.6, 3.2$\}$.

%

We also allow $d_1$ to vary over values between 20 cm and 380 cm, and the return separation $d_2-d_1$ to vary between 40 cm and 250 cm; and each instance is generated with many noise vectors.  In total, we generate 261,000 examples.

\begin{figure}[!t]
\centerline{\hbox{
\begin{tabular}{cc}
\picwh{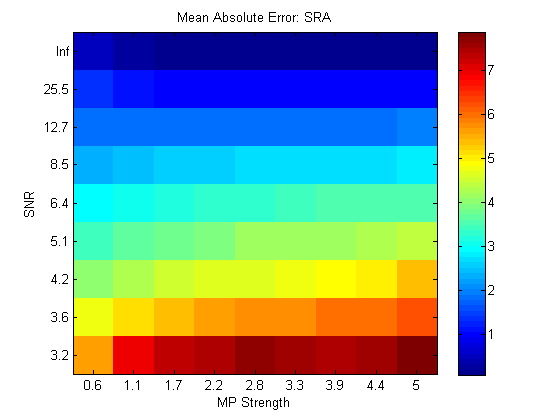}{0.35}{0.14} & \picwh{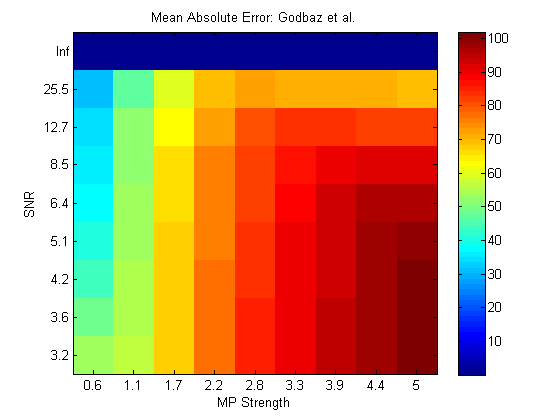}{0.35}{0.14} 
\end{tabular}
}}
\caption{Two-Path Simulation: Mean Absolute Error (cm).  Left: SRA.  Right: Godbaz.}
\label{fig:TwoPath}
\end{figure}

We visualize the results in Figure \ref{fig:TwoPath}, in which we show the mean absolute error (MAE) of the depth estimates as a function of multipath strength and SNR.  Each square corresponds to a (multipath strength, SNR) pair; we average over all examples falling into the square to compute the MAE.

\para{Discussion}  In general, SRA's behavior is as we would imagine: as multipath strength increases, MAE increases; likewise, as SNR decreases, MAE increases.

There are two notable facts about the results.  First, the MAE is quite small for ``realistic'' values of multipath strength and SNR.  Focus on the upper left rectangle of Figure \ref{fig:TwoPath} consisting of SNRs from $\infty$ to $8.5$ and multipath strengths from $0.6$ to $2.2$; note that these are still quite challenging values.  In this regime, the MAE is low: it is less than 2.6 cm in all squares, with an average of 1.4 cm.

Second, SRA's performance degrades in a graceful way: the MAE increases gradually in both dimensions.  In fact, even when the multipath is 5 times stronger than the true return and the SNR has a low value of 3.2, the MAE is 7.9 cm, which is a very reasonable error in such circumstances.

It is interesting to compare SRA's MAE with that of Godbaz \etal \cite{godbaz2012closed}, recalling that Godbaz's algorithm was designed with two-path multipath in mind.  Godbaz gives good results when there is no noise: the MAE is nearly 0.  Once even a small amount of noise is added in, however, Godbaz's performance drops significantly.  For example, with SNR = 25.5, the errors range from 30.8 cm (MP Strength = 0.6) to 69.7 cm (MP Strength = 5).  Performance worsens significantly as SNR decreases.


\begin{figure}[!t]
\centerline{\hbox{
\pic{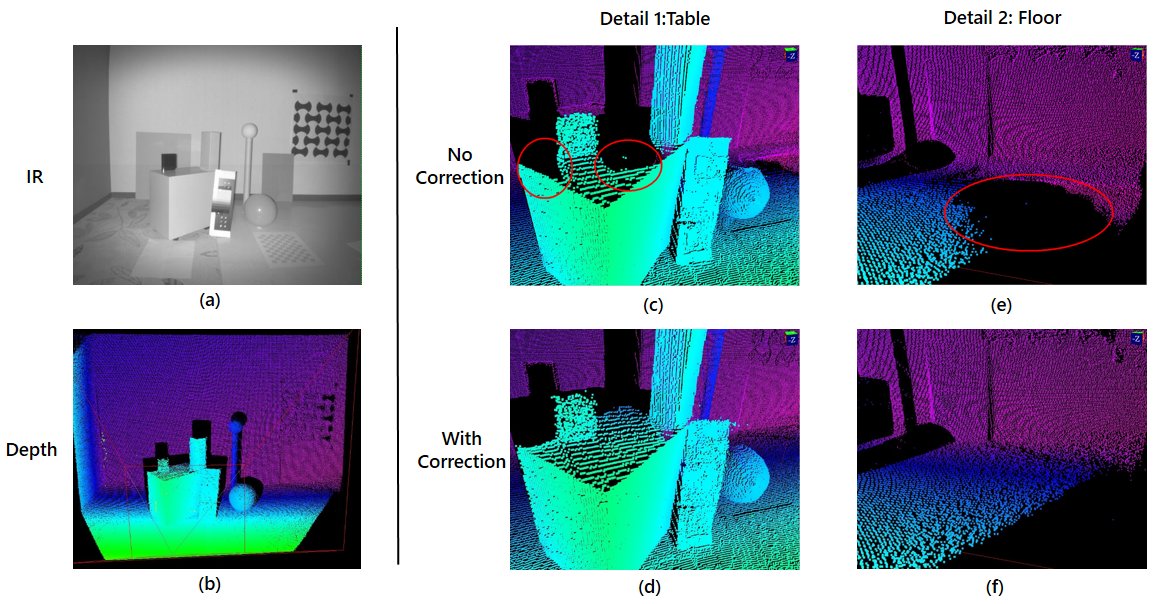}{0.8}   
}}
\caption{``Geometric''. (a) IR image.  (b) SRA depth estimate. (c,e) Optimal depth estimates without multipath correction; errors are circled in red.  (d,f) SRA depth.\vs}
\label{fig:Geometric}
\end{figure}

\subsection{Real Images}
\label{sec:RealImages}

We now discuss the results of running SRA on three different challenging images.  The images are collected using the Xbox One sensor, which has $m=3$ modulation frequencies and a resolution of $424 \times 512$.

Unfortunately, we are not able to compare with any of the real-time competing methods due to their incompatibility with the Xbox One imaging setup. Specifically, Godbaz \etal \cite{godbaz2012closed} and Dorrington \etal \cite{dorrington2011separating} require a very particular relation between the modulation frequencies, which is not satisfied by the Xbox One's modulation frequencies.  Kirmani \etal \cite{kirmani2013spumic} requires 5 modulation frequencies, while a second method described in \cite{godbaz2012closed} requires 4; the Xbox One uses only 3 modulation frequencies.  Thus, for comparison purposes in this section, we run SRA against a variant of SRA which looks for the optimal single path which best describes the sensor measurement, which we call ``Opt-Single''.

\para{``Strips'' Image}  See Figure \ref{fig:Teaser}. This image is a simple scene -- a floor and a wall; however, the floor is composed of strips of different materials, each of which has different reflectance properties.  This can be seen clearly in the IR image in Figure \ref{fig:Teaser}.

The more specular materials tend to lead to very high multipath: a path which bounces off the wall, and from there to the floor, will generally have strength higher than the direct return from the floor.  This is due to the fact that the direct path is nearly parallel to the floor, leading to a weak direct return.  

Results of the depth reconstruction are shown in Figure \ref{fig:Teaser}.  To see the effect of multipath removal, we compare SRA's depth estimate to Opt-Single.  In Figure \ref{fig:Teaser}, one can see details of the scene which show the effect of ``specular floor'' multipath: the wall is effectively reflected into the floor, leading to grossly inaccurate estimates for the floor.  By contrast, SRA reconstructs a very clean floor, which is seen to be almost completely flat, the exceptions being a few small depressions of less than 4 cm.

Note in Figure \ref{fig:Teaser} that SRA invalidates a number of pixels at the top right part of the image, corresponding to a patterned wall-hanging; this is due to the low reflectivity of this part of the scene, leading to very noisy measurements.

\para{``Geometric'' Image} See Figure \ref{fig:Geometric}. This image consists of a similar set-up to ``Strips'', but with various geometric objects inserted.  These objects include many sharp angles, as well as several thin structures.  SRA reconstructs the scene quite well, see Figure \ref{fig:Geometric}.  Again, note the scene details in Figure \ref{fig:Geometric}, which show the performance of SRA vs. Opt-Single.  Much of the surface of the rectangular table is lost without accounting for multipath, whereas SRA is able to reconstruct it.  And the corner where the wall joins the floor is lost -- actually reflected into the floor -- without accounting for multipath, whereas SRA recovers it.

\begin{figure}[!t]
\centerline{\hbox{
\pic{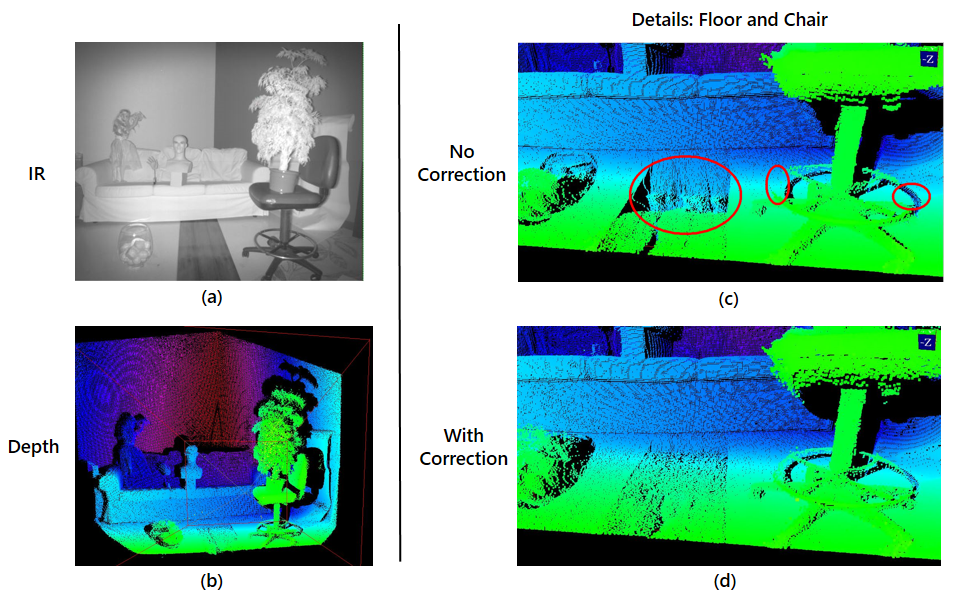}{0.6}  
}}
\caption{``Living Room''. (a) IR image.  (b) SRA depth estimate. (c) Optimal depth estimate without multipath correction; errors are circled in red.  (d) SRA depth estimate.\vs}
\label{fig:LivingRoom}
\end{figure}

\para{``Living Room'' Image} See Figure \ref{fig:LivingRoom}. This image consists of a couch as well as a number of objects inserted.  As in ``Strips'', there is a strip of reflective material on the floor; Opt-Single has trouble accounting for this strip, as can be seen in Figure \ref{fig:LivingRoom}.  Perhaps more interestingly, the thin structure which is the wheel at the base of the swivel chair is largely reconstructed by SRA, whereas large portions of it, on both the right and left sides, are lost by Opt-Single.

\section{Conclusions}
\label{sec:conclusions}

We have presented the SRA algorithm for removing multipath interference from ToF images.  We have seen that the method is both general, dealing with many types of multipath, and fast.  SRA has been experimentally validated on both synthetic data as well as challenging real images, demonstrating its superior performance.



\bibliographystyle{splncs}
\bibliography{Paper}

\begin{thebibliography}{10}

\bibitem{han2013enhanced}
Han, J., Shao, L., Xu, D., Shotton, J.:
\newblock Enhanced computer vision with {Microsoft Kinect} sensor: A review.
\newblock IEEE Transactions on Cybernetics \textbf{43}(5) (2013)  1318--1334

\bibitem{xbox2014xboxone}
Microsoft:
\newblock {Xbox One Sensor}.
\newblock \url{http://www.xbox.com/en-US/xbox-one/innovation}

\bibitem{falie2008distance}
Falie, D., Buzuloiu, V.:
\newblock Distance errors correction for the time of flight ({ToF}) cameras.
\newblock In: Imaging Systems and Techniques, 2008. IST 2008. IEEE
  International Workshop on, IEEE (2008)  123--126

\bibitem{falie2008further}
Falie, D., Buzuloiu, V.:
\newblock Further investigations on {ToF} cameras distance errors and their
  corrections.
\newblock In: European Conference on Circuits and Systems for Communications
  (ECCSC). (2008)  197--200

\bibitem{fuchs2010multipath}
Fuchs, S.:
\newblock Multipath interference compensation in time-of-flight camera images.
\newblock In: International Conference on Pattern Recognition (ICPR). (2010)
  3583--3586

\bibitem{fuchs2013compensation}
Fuchs, S., Suppa, M., Hellwich, O.:
\newblock Compensation for multipath in {ToF} camera measurements supported by
  photometric calibration and environment integration.
\newblock In: Computer Vision Systems.
\newblock (2013)  31--41

\bibitem{jimenez2012modelling}
Jim{\'e}nez, D., Pizarro, D., Mazo, M., Palazuelos, S.:
\newblock Modelling and correction of multipath interference in time of flight
  cameras.
\newblock In: IEEE Conference on Computer Vision and Pattern Recognition
  (CVPR). (2012)  893--900

\bibitem{dorrington2011separating}
Dorrington, A.A., Godbaz, J.P., Cree, M.J., Payne, A.D., Streeter, L.V.:
\newblock Separating true range measurements from multi-path and scattering
  interference in commercial range cameras.
\newblock In: IS\&T/SPIE Electronic Imaging, International Society for Optics
  and Photonics (2011)  786404--786404

\bibitem{godbaz2012closed}
Godbaz, J.P., Cree, M.J., Dorrington, A.A.:
\newblock Closed-form inverses for the mixed pixel/multipath interference
  problem in amcw lidar.
\newblock In: IS\&T/SPIE Electronic Imaging, International Society for Optics
  and Photonics (2012)  829618--829618

\bibitem{kirmani2013spumic}
Kirmani, A., Benedetti, A., Chou, P.A.:
\newblock Spumic: Simultaneous phase unwrapping and multipath interference
  cancellation in time-of-flight cameras using spectral methods.
\newblock In: IEEE International Conference on Multimedia and Expo (ICME).
  (2013)  1--6

\bibitem{candes2005decoding}
Candes, E.J., Tao, T.:
\newblock Decoding by linear programming.
\newblock IEEE Transactions on Information Theory \textbf{51}(12) (2005)
  4203--4215

\bibitem{candes2006stable}
Candes, E.J., Romberg, J.K., Tao, T.:
\newblock Stable signal recovery from incomplete and inaccurate measurements.
\newblock Communications on Pure and Applied Mathematics \textbf{59}(8) (2006)
  1207--1223

\bibitem{donoho2006compressed}
Donoho, D.L.:
\newblock Compressed sensing.
\newblock IEEE Transactions on Information Theory \textbf{52}(4) (2006)
  1289--1306

\bibitem{wright2009robust}
Wright, J., Yang, A.Y., Ganesh, A., Sastry, S.S., Ma, Y.:
\newblock Robust face recognition via sparse representation.
\newblock IEEE Transactions on Pattern Analysis and Machine Intelligence
  \textbf{31}(2) (2009)  210--227

\end{thebibliography}

\end{document}